\documentclass[runningheads]{llncs}
\usepackage{amssymb}
\usepackage{graphicx,verbatim}
\usepackage{booktabs}
\usepackage{mathtools}
\usepackage{hyperref}
\usepackage{pifont}
\usepackage{amsmath}
\usepackage{multirow}
\usepackage[T1]{fontenc}
\usepackage{graphicx,verbatim}
\begin{document}
\title{Retrieving Patient-Specific Radiomic Feature Sets for Transparent Knee MRI Assessment}
\titlerunning{Retrieving Radiomic Feature Sets for Knee MRI}
%

\author{Yaxi Chen\inst{1,2}\orcidID{0009-0007-5906-899X} 
\and Simin Ni\inst{3}\orcidID{0009-0007-2780-6118} 
\and Jingjing Zhang\inst{1}\orcidID{0009-0005-4808-8413} 
\and Shaheer U. Saeed\inst{2,4,5,6}\orcidID{0000-0002-5004-0663} 
\and Yipei Wang\inst{2}\orcidID{0000-0002-9589-7177} 
\and Aleksandra Ivanova\inst{3}\orcidID{0009-0000-4113-8928} 
\and Rikin Hargunani\inst{7}\orcidID{0000-0002-0953-8443} 
\and Chaozong Liu\inst{3,7}\orcidID{0000-0002-9854-4043} 
\and Jie Huang\inst{1}\orcidID{0000-0001-7951-2217} 
\and Yipeng Hu\inst{2,4}\orcidID{0000-0003-4902-0486}
}

\authorrunning{Y. Chen et al.}

%

\institute{Mechanical Engineering Department, University College London, London, UK \and Hawkes Institute, University College London, London, UK \and Institute of Orthopaedic \& Musculoskeletal Science, University College London, Royal National Orthopaedic Hospital, Stanmore, UK \and School of Engineering and Materials Science, Queen Mary University of London, London, UK \and Centre for Bioengineering, Queen Mary University of London, London, UK \and Department of Medical Physics and Biomedical Engineering, University College London, London, UK \and
Royal National Orthopaedic Hospital, Stanmore, UK}
  
\maketitle             
\begin{abstract}
In knee MRI analysis, classical radiomic features are designed to quantify image appearance and intensity patterns. Compared with end-to-end deep learning (DL) models trained for disease classification, radiomics pipelines with low-dimensional parametric classifiers offer enhanced transparency and interpretability, yet often underperform because of the reliance on population-level predefined feature sets. Recent work on adaptive radiomics uses DL to predict feature weights over a large radiomic pool, then thresholds these weights to retain the ``top-$k$'' features from $\mathcal{F}$ (often $F=\lvert\mathcal{F}\rvert\sim 10^3$). However, such marginal ranking can over-admit redundant descriptors and overlook complementary feature interactions. We propose a patient-specific feature-set selection framework that predicts a single compact feature set per subject, targeting complementary and diverse evidence rather than marginal top-k features. To overcome the intractable combinatorial search space of $\binom{F}{k}$ features, our method utilizes a two-stage retrieval strategy (i) randomly sample diverse candidate feature sets, then (ii) rank these sets with a learned scoring function to select a high-performing ``top-$1$'' feature set for the specific patient. The system consists of a feature-set scoring function, and a downstream classifier that performs the final diagnosis using the selected set. We empirically show that the proposed two-stage retrieval approximates the original exhaustive $\binom{F}{k}$ selection, with a 95$^{th}$-percentil error of $0.0055$. Validating on tasks including ACL tear detection and Kellgren-Lawrence grading for osteoarthritis, the experimental results achieve diagnostic performance, outperforming the top-$k$ approach with the same $k$ values, and competitive with end-to-end DL models while maintaining high transparency. The model generates auditable feature sets that link clinical outcomes to specific anatomical regions and radiomic families, allowing clinicians to inspect which anatomical structures and quantitative descriptors drive the prediction. The codes are available at: \url{https://github.com/YaxiiC/OA_KLG_Retrieval.git}

\end{abstract}

\section{Introduction}
\label{sec:intro}

Knee MRI is increasingly central to the assessment of both traumatic and degenerative knee disorders because it provides noninvasive, whole-joint visualisation of soft-tissue structures, including the ligaments, menisci, and articular cartilage, that directly informs diagnosis and treatment planning~\cite{nguyen2014mr,wu2022preoperative,ehmig2023mr}. By enabling quantitative evaluation of tissue morphology and heterogeneity beyond visual inspection, MRI supports imaging biomarker development and motivates unified computational frameworks for tasks such as osteoarthritis (OA) grading and anterior cruciate ligament (ACL) injury classification, where grading reliability and tear subtype differentiation remain clinically challenging~\cite{yao2024cartimorph,damen2014inter,stone2021management,hannon2024american}. Radiomics extends MRI by converting routine scans into quantitative features that can capture subtle pathological changes~\cite{rogers2020radiomics}. However, conventional radiomics pipelines often rely on predefined, fixed feature sets, which have been shown over-restrictive for inter-subject variability and sensitive to data heterogeneity, motivating more adaptive feature-selection strategies~\cite{chen2025patient,chen2025radiomic}.


\noindent End-to-end DL models learn task-specific representations directly from images and often achieve strong predictive performance, but their decision mechanisms are typically difficult to explain beyond coarse saliency visualisations or resorting to additionally-required concept activation. To retain interpretability while improving flexibility, recent adaptive radiomics methods use an image-conditioned selector to predict per-feature relevance scores over a large radiomic pool, followed by weighting and thresholding to obtain a ``top-$k$'' subset for downstream prediction~\cite{chen2025patient,chen2025radiomic}. However, this feature-wise selection treats features conditionally independent: it does not optimally account for redundancy and complementarity among radiomic descriptors, and may over-admit correlated features or miss conditional interactions. From a combinatorial perspective, marginal screening selects among $F$ singletons (i.e., $\binom{F}{1}$), whereas a compact explanation should ideally be a fixed-size set of $k$ features chosen from the $\binom{F}{k}$ possible sets.

\noindent In this study, we propose a patient-specific feature-set selection framework for interpretable knee MRI assessment. For each subject, the explanation takes the form of an explicit set of predefined size $k$, where each selected feature is traceable to its anatomical ROI and radiomic family. Rather than attempting an intractable exhaustive search, we use a computationally feasible retrieval-based procedure that approximates set selection sufficiently well in practice, yielding substantially stronger performance than conventional top-$k$ selection. During training, a lightweight radiomics-only probe provides a reward signal to learn set utility, while a downstream classifier performs the final diagnosis using only the retrieved set. We validate the approach on multiple real-world knee MRI tasks, including KL osteoarthritis grading and ACL tear detection.

\noindent In summary, we make three contributions: (1) a patient-specific radiomic feature set that produces compact, auditable explanations with ROI- and family-level traceability; (2) a two-stage retrieval strategy for feature-set selection, comprising random exploration to initialise the scoring function in Stage~1 and large-pool sampling followed by candidate pooling and a top set retrieval in Stage~2, enabling feasible optimisation over the combinatorial space; and
(3) a scoring function coupled with a downstream classifier that surpasses top-k selection at fixed k, achieves competitive performance relative to end-to-end DL while retaining interpretability, and generalises across multiple real-world clinical tasks.

\section{Methods}
\label{sec:method}

\noindent\textbf{Problem formulation.} Let $\mathcal{D}=\{(\mathbf{x}_i,y_i)\}_{i=1}^{N}$ denote the dataset, where $\mathbf{x}_i\in \mathcal{X}=\mathbb{R}^{D\times H\times W}$ is the 3D knee MRI volume of subject $i$ and $y_i\in \mathcal{Y}=\{1,\dots,C\}$ is its label.
Let $a\in\mathcal{A}$ index the set of anatomical ROIs (e.g., femur, cartilage), and
let $m\in\mathcal{M}=\{0,1\}^{|\mathcal{A}|\times D\times H\times W}$ denote the multi-ROI mask space.
A pre-trained segmentation model $g:\mathcal{X}\rightarrow\mathcal{M}$ produces ROI masks
$\mathbf{m}_{i}=g(\mathbf{x}_i)$, with $\mathbf{m}_{i,a}\in\{0,1\}^{D\times H\times W}$ the mask for ROI $a\in\mathcal{A}$.
The radiomic feature extractor $e:\mathcal{X}\times\mathcal{M}\rightarrow \mathcal{R}$ maps an image and its ROI masks to a pooled radiomics vector: 
\[\mathbf{r}_i=e(\mathbf{x}_i,\mathbf{m}_i)\in\mathcal{R}=\mathbb{R}^{F},
\quad \mathcal{F}=\{1,\dots,F\},
\]
Here, $\mathcal{F}$ denotes the feature index pool of size $F$, and $f\in\mathcal{F}$ indexes the $f$-th scalar entry of $\mathbf{r}_i$; we use $r_{i,f}$ to denote its value.

\begin{figure}
    \centering
    \includegraphics[width=0.85\linewidth]{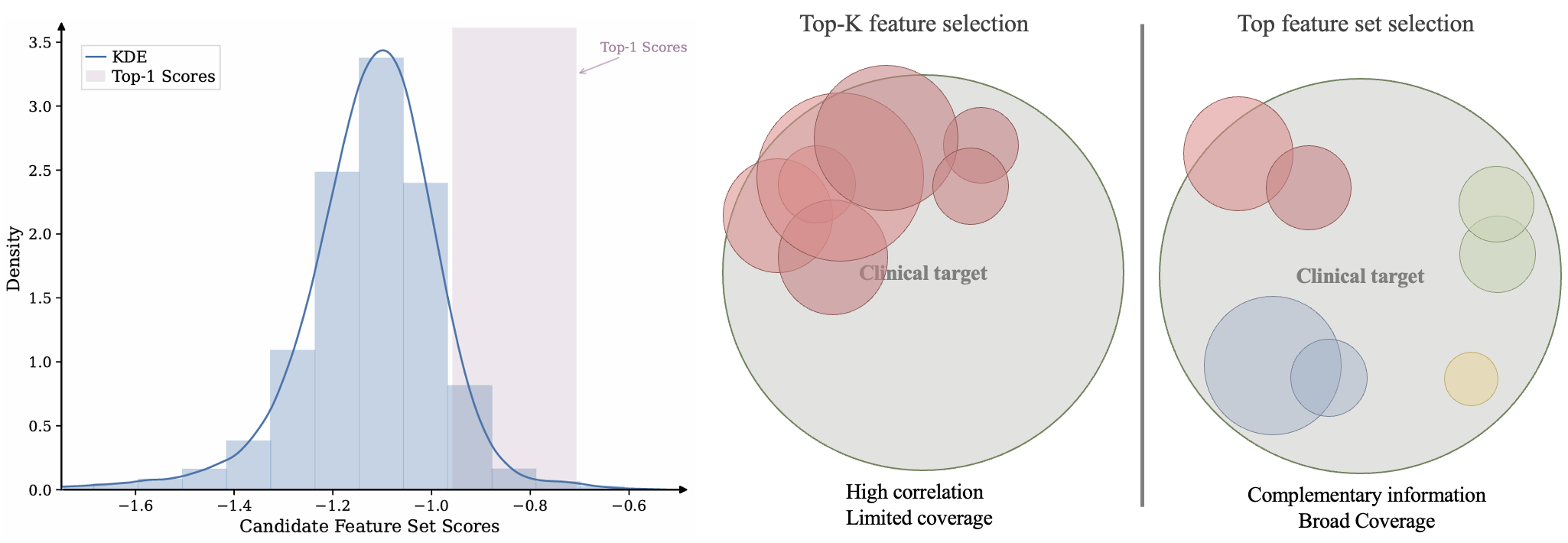}
    \caption{\textbf{Motivation and feasibility of feature-set retrieval.}
\textbf{Left:} Candidate feature-set score distribution (blue histogram) estimated via kernel density estimation (KDE); the red-shaded band marks the top-1 score range, illustrating near-optimal set identification without enumerating.
\textbf{Right:} Illustration of top-$k$ vs. proposed feature-set selection approaches, where top-$k$ may select correlated features (limited coverage), whereas feature-set selection provides complementary evidence (broader coverage).}
    \label{fig:histo}
\end{figure}

We consider fixed-size feature sets drawn from
\(
\binom{\mathcal{F}}{k}:=\{S\subseteq \mathcal{F}\,:\,|S|=k\},
\)
i.e., the set of all possible \(k\)-feature combinations from the pool \(\mathcal{F}\), enforcing a constant, auditable feature budget. Here \(\binom{\mathcal{F}}{k}\) denotes this combinatorial family of candidate sets, while \(\binom{F}{k}\) gives the number of possible \(k\)-feature sets. 
$S$ denotes a set of feature indices (a $k$-subset of $\mathcal{F}$). For \(S\in\binom{\mathcal{F}}{k}\), each selected index \(f\in S\) is associated with \(r_{i,f}\) and feature metadata (e.g., ROI, feature family, and type).

\noindent\textbf{Motivation.}
For each subject $i$, our goal is to identify a patient-specific set of fixed cardinality $k$, i.e., a feature set $S_i^\star\in\binom{\mathcal{F}}{k}$, so that the final prediction is attributable to an explicit list of $k$ radiomic features.
Because exhaustively optimising over $\binom{F}{k}$ is infeasible (in our case, $\sim 10^{55}$), regardless the problem is defined as a $\binom{F}{k}$-class classification or as a direct $\binom{F}{k}$ content retrieval. we approximate $S_i^\star$ using a two-stage retrieval strategy that reduces set selection from a $\binom{F}{k}$-way search to a $\binom{M}{1}$ choice, where $M$ denotes the size of a small per-subject candidate pool constructed in Stage~2.

\noindent\textbf{Feature set encoding and scoring function.} For a candidate index set $S\in\binom{\mathcal{F}}{k}$, let
\(
\mathbf{r}_{i,S}:=(r_{i,f})_{f\in S}\in\mathbb{R}^{k}
\)
denote the radiomics subvector of subject $i$ indexed by $S$.
A shared permutation-invariant set encoder $\mathcal{E}_{\theta}$, parameterised by weights $\theta$, tokenises each selected feature (value + metadata) and mean-pools the $k$ tokens to obtain a $d$-dimensional set embedding
\(
\mathbf{z}_i(S):=\mathcal{E}_{\theta}\!\big(\mathbf{r}_{i,S},S\big).
\) We then learn an MRI-conditioned scorer $s_{\psi}:\mathcal{X}\times\mathbb{R}^{d}\rightarrow\mathbb{R}$ , parameterised by weights $\psi$, that assigns a utility score to $S$ for subject $i$:
\(
s_{\psi}\!\big(\mathbf{x}_i,\mathbf{z}_i(S)\big)\in\mathbb{R}.
\)
Higher scores indicate more predictive evidence sets for diagnosis. Candidate sets are generated by the two-stage retrieval procedure below.

\noindent\textbf{Probe reward.} Ground-truth optimal feature sets are unobserved, and direct supervision of set selection from the downstream classifier is sparse and unstable under top-1 decisions over a combinatorial space. We therefore use a training-only linear probe reward as a dense, low-variance proxy that approximates the rapid separability of a candidate set ~\cite{shi2022efficacy,chen2020simple}. To compute the reward, we draw two disjoint subsets from the training data: a support set $\mathcal{D}_{\mathrm{sup}}$ used to fit the probe, and a query set $\mathcal{D}_{\mathrm{qry}}$ used to evaluate it.
We fit a lightweight linear probe $q_{\phi}:\mathbb{R}^{k}\rightarrow\mathbb{R}^{C}$ (parameters $\phi$) on $\mathcal{D}_{\mathrm{sup}}$ using $\{(\mathbf{r}_{u,S},y_u)\}_{u\in\mathcal{D}_{\mathrm{sup}}}$, and evaluate it on $\mathcal{D}_{\mathrm{qry}}$ to obtain the average cross-entropy $\ell_{\mathrm{qry}}(S)$. We then define the probe reward as
\(
R(S)=-\,\ell_{\mathrm{qry}}(S),
\) so that larger rewards correspond to more predictive candidate sets.
Patient-specificity is preserved because the ranking is conditioned on $\mathbf{x}_i$ through $s_{\psi}(\mathbf{x}_i,\mathbf{z}_i(S))$, while the probe reward provides a complementary population-level signal reflecting shared disease-related patterns.

\begin{figure}
    \centering
    \includegraphics[width=0.86\linewidth]{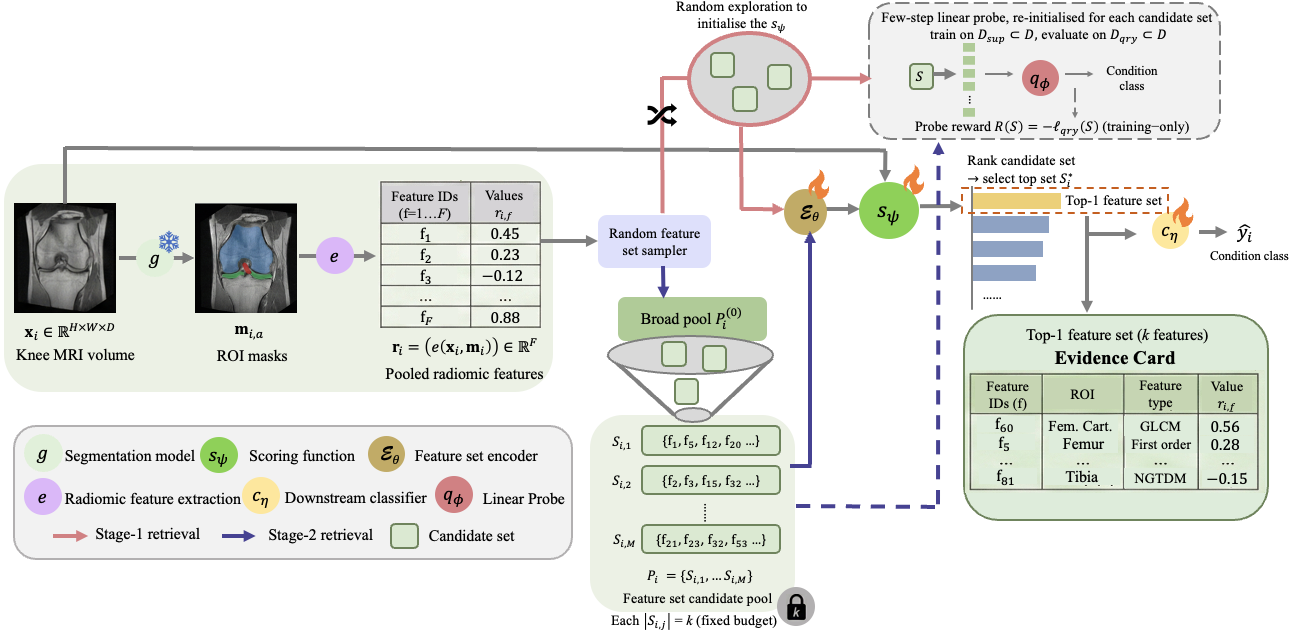}
    \caption{Overview of the proposed feature set selection framework. The red arrow indicates the stage 1 retrieval and the blue arrow indicates the stage 2 retrieval.}
    \label{fig:placeholder}
\end{figure}
\noindent\textbf{Two-stage retrieval strategy.} \emph{\textbf{Stage~1 (random exploration):}} During early training, for each subject $i$ we sample candidate sets $S\sim\binom{\mathcal{F}}{k}$ at random and use their probe rewards to train the scoring function. This broad exploration provides diverse supervision and stabilises learning of an initial scoring function. \emph{\textbf{Stage~2 (top set retrieval):}} For each subject $i$, we first sample a large preliminary pool  $\mathcal{P}^{(0)}_i\subset\binom{\mathcal{F}}{k}$, score all candidates with the current scoring function, and retain the top $M$ sets to form the candidate feature set pool:
\[
\mathcal{P}_i:=\{S_{i,1},\dots,S_{i,M}\}\subset\mathcal{P}^{(0)}_i,\qquad M\ll |\mathcal{P}^{(0)}_i|.
\]
We then select the top-ranked set for prediction,
\[
S_i^\star = S_{i,j^\star},\qquad 
j^\star=\arg\max_{j\in\{1,\dots,M\}} s_{\psi}\!\big(\mathbf{x}_i,\mathbf{z}_i(S_{i,j})\big),\qquad |S_i^\star|=k,
\]
and feed $S_i^\star$ to the downstream classifier.
We sample a small subset of candidates from the current pool,
$\mathcal{J}_i\subseteq\{1,\dots,M\}$ with $|\mathcal{J}_i|=Q$,
compute their probe rewards to provide ongoing supervision for $s_\psi$.

In theory, the above two stages could be iterated to further approximating the ``true'' $\binom{F}{k}$ combinatorial optimality, however a single alternation was found to provide sufficient performance gain reported in this work.

\noindent\textbf{Retrieval gap.}
For subject $i$, let the infeasible optimal reward be denoted as \(
R_i^{\max}:=\max_{S\in\binom{\mathcal{F}}{k}} R_i(S),
\) where $R_i(S)$ denotes the reward for features set $S$.
The subject-level retrieval gap
\(
\Delta_i := R_i^{\max}-R_i(S_i^\star)\;\ge 0,
\)
where $S_i^\star$ is the retrieved top-1 set under our proposed two-stage retrieval strategy.
The expected gap over subjects admits the tail-integral identity
\(
\mathbb{E}_i[\Delta_i]
=
\int_{0}^{\infty}\mathbb{P}_i\!\left(\Delta_i>\epsilon\right)\,d\epsilon,
\)
i.e., the average retrieval gap equals the area under the tail-probability curve of $\Delta_i$.
Since computing $R_i^{\max}$ requires an intractable exhaustive search over $\binom{F}{k}$ feature sets, it is computationally infeasible to estimate this quantity directly; instead, we report the empirical distribution of the achieved top-1 rewards $R_i(S_i^\star)$ across subjects (Fig.~\ref{fig:histo}, left) as a practical indicator of retrieval quality under a fixed search budget.
Although candidate rewards exhibit approximately Gaussian variability, the selected subsets consistently occupy the extreme right tail, suggesting effective discrimination among high-performing candidate sets. This efficacy is consistent with the known similarity and correlation between groups of radiomic features.

\noindent\textbf{Downstream classifier.}
Given the selected top-1 evidence set $S_i^\star$, we compute its permutation-invariant set representation using the shared encoder,
\(
\mathbf{z}_i^\star := \mathbf{z}_i(S_i^\star)=\mathcal{E}_{\theta}\!\big(\mathbf{r}_{i,S_i^\star},S_i^\star\big).
\)
A downstream classifier $c_{\eta}:\mathbb{R}^{d}\rightarrow\mathbb{R}^{C}$ (parameters $\eta$) maps $\mathbf{z}_i^\star$ to class logits
\(
\boldsymbol{\ell}_i=c_{\eta}(\mathbf{z}_i^\star)\in\mathbb{R}^{C},
\)
and predicts class probabilities via
\(
p_{\eta}(y_i\mid S_i^\star)
=\mathrm{softmax}\!\big(\boldsymbol{\ell}_i\big)_{y_i}.
\)
Notably, $p_{\eta}(y_i\mid S_i^\star)$ depends on $\mathbf{x}_i$ only through the explicit, auditable evidence specified by $S_i^\star$.

\noindent\textbf{Learning objective.}
We jointly learn the scorer $s_{\psi}$ and the downstream classifier $c_{\eta}$.
For each subject $i$, we compute a task-specific classification loss $\mathcal{L}_{\mathrm{cls}}(i)$ from the logits $\boldsymbol{\ell}_i$ obtained using the selected set $S_i^\star$.
To supervise the scorer, we evaluate probe rewards on a subset of candidates indexed by $\mathcal{J}_i\subseteq\{1,\dots,M\}$ with $|\mathcal{J}_i|=Q$ and regress scores to rewards:
\[
\mathcal{L}_{\mathrm{scr}}(i)
=
\frac{1}{Q}\sum_{j\in\mathcal{J}_i}
\Big(s_{\psi}\!\big(\mathbf{x}_i,\mathbf{z}_i(S_{i,j})\big)-R(S_{i,j})\Big)^2.
\]

\begin{figure}[t]
    \centering
    \includegraphics[width=0.88\linewidth]{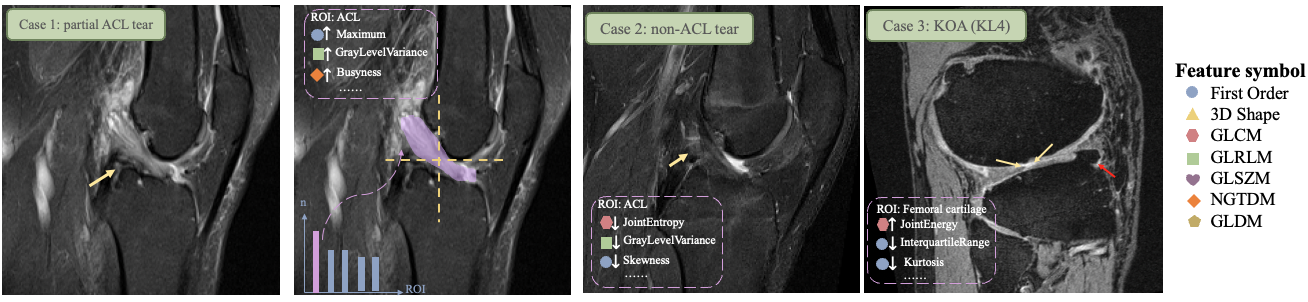}
    \caption{\textbf{Case studies for model interpretation.} Examples of (Case 1) partial ACL tear, (Case 2) non-tear, and (Case 3) KL4 knee osteoarthritis. The histogram shows, for each ROI, the number of selected features contributing to the prediction. The symbols and their adjacent arrows indicate whether the corresponding feature values are relatively high or low compared with the rest of the dataset (z-score).}
    \label{fig:case_study}
\end{figure}
\noindent Final objective: \(\mathcal{L}(i)=\mathcal{L}_{\mathrm{cls}}(i)+\lambda_{\mathrm{scr}}\,\mathcal{L}_{\mathrm{scr}}(i),\;(\psi^\star,\theta^\star,\eta^\star)=\arg\min_{\psi,\theta,\eta}\sum_{i=1}^{N}\mathcal{L}(i).\)

\section{Experiments and Results}
\begin{table}[!ht]
\renewcommand{\thetable}{1}
\centering
\caption{Comparison of different configurations and methods. k: the feature set size; Pstep: number of probe gradient steps; Npool: size of the candidate pool in stage-2; Ens: the use of ensembling during inference; RS: random sampling; OA-5: 5-class classification; OA-B: binary classification. We \textbf{bold} the top-2 Acc/BAcc scores. }
\label{ablation}
\renewcommand{\arraystretch}{0.85}
\setlength{\tabcolsep}{0.6pt} 
\fontsize{8pt}{8pt}\selectfont


\begin{tabular}{|c|c|c|c|c|c|ccccc|}
\hline
 &  &  &  &  &  & \multicolumn{5}{c|}{\textbf{Evaluation Metrics}} \\
\cline{7-11}
\textbf{Method} & \textbf{k} & \textbf{Pstep} & \textbf{Npool} & \textbf{Ens} & \textbf{Task} &
\textbf{Acc.} & \textbf{MacroF1} & \textbf{BAcc} & \textbf{AUC} & \textbf{QWK} \\
\hline

\textbf{\multirow{3}{*}{Pstep\(\downarrow\)}} & \multirow{3}{*}{30} & \multirow{3}{*}{5} & \multirow{3}{*}{1000} & \multirow{3}{*}{3}
& ACL &0.59±0.05& 0.36±0.06& 0.38±0.07 & 0.59±0.02 & 0.18±0.03\\
&&&&& OA-b &0.69±0.16& 0.65±0.17 & 0.66±0.17 & 0.80±0.06 & - \\
&&&&& OA-5 &0.37±0.02&0.26±0.04& 0.31±0.03 & 0.66±0.03 & 0.30±0.06 \\
\hline

\textbf{\multirow{3}{*}{k=20}} & \multirow{3}{*}{20} & \multirow{3}{*}{10} & \multirow{3}{*}{1000} & \multirow{3}{*}{3}
& ACL &0.67±0.03& 0.48±0.07& 0.46±0.06 &  0.65±0.03 & 0.34±0.07\\
&&&&& OA-b &0.74±0.05& 0.69±0.08 & 0.68±0.07 & 0.71±0.20 & - \\
&&&&& OA-5 &0.37±0.02& 0.27±0.04 & 0.31±0.03 & 0.66±0.03 & 0.30±0.05 \\
\hline

\textbf{\multirow{3}{*}{k=50}} & \multirow{3}{*}{50} & \multirow{3}{*}{10} & \multirow{3}{*}{1000} & \multirow{3}{*}{3}
& ACL &\textbf{0.75±0.01}& 0.32±0.03 & 0.35±0.01 & 0.68±0.02 & 0.04±0.04 \\
&&&&& OA-b &0.74±0.08& 0.72±0.09 & \textbf{0.72±0.08} & 0.80±0.13 & - \\
&&&&& OA-5 &\textbf{0.47±0.06}& 0.39±0.03 & 0.41±0.05 & 0.82±0.09 & 0.54±0.22 \\
\hline

\textbf{\multirow{3}{*}{Npool\(\downarrow\)}} & \multirow{3}{*}{30} & \multirow{3}{*}{10} & \multirow{3}{*}{500} & \multirow{3}{*}{3}
& ACL &0.65±0.02& 0.38±0.03& 0.39±0.04& 0.60±0.04 & 0.22±0.08 \\
&&&&& OA-b &0.69±0.10 &0.65±0.10 & 0.66±0.10 & 0.80±0.05 & - \\
&&&&& OA-5 &0.34±0.06& 0.22±0.05 & 0.28±0.05 & 0.65±0.10 & 0.24±0.11 \\
\hline

\textbf{\multirow{3}{*}{RS}} & \multirow{3}{*}{30} & \multirow{3}{*}{10} & \multirow{3}{*}{-} & \multirow{3}{*}{3}
& ACL &0.64±0.04 & 0.48±0.07 &0.47±0.06  & 0.64±0.03 & 0.34±0.07\\
&&&&& OA-b & 0.70±0.08 & 0.69±0.07&0.70±0.05& 0.69±0.07 &  -  \\
&&&&& OA-5 &0.33±0.09& 0.31±0.08& 0.36±0.06&0.67±0.03 & 0.42±0.13 \\
\hline

\textbf{\multirow{3}{*}{NoEns}} & \multirow{3}{*}{30} & \multirow{3}{*}{10} & \multirow{3}{*}{1000} & \multirow{3}{*}{1}
& ACL & 0.63±0.05 & 0.37±0.02 & 0.39±0.03 & 0.58±0.08 & 0.20±0.10 \\
&&&&& OA-b &0.68±0.09& 0.69±0.08 & 0.71±0.09 & 0.78±0.1 & - \\
&&&&& OA-5 &0.39±0.06 &0.37±0.07 & 0.40±0.06& 0.67±0.09& 0.49±0.15 \\
\hline

\textbf{\multirow{3}{*}{E2E}} & \multirow{3}{*}{-} & \multirow{3}{*}{-} & \multirow{3}{*}{-} & \multirow{3}{*}{-}
& ACL &0.70±0.05& 0.37±0.06& 0.38±0.05 &  0.57±0.12 &  0.22±0.14 \\
&&&&& OA-b &\textbf{0.76±0.08}& 0.71±0.10 & 0.70±0.09 & 0.83±0.04 & - \\
&&&&& OA-5 &0.41±0.08&0.36±0.07 & 0.38±0.06 & 0.74±0.07 & 0.58±0.14 \\
\hline

\textbf{\multirow{3}{*}{img+msk}} & \multirow{3}{*}{-} & \multirow{3}{*}{-} & \multirow{3}{*}{-} & \multirow{3}{*}{-}
& ACL & 0.72±0.03 &  0.48±0.13 & 0.47±0.11 & 0.74±0.05 & 0.35±0.25 \\
&&&&& OA-b & 0.70±0.11& 0.69±0.10 & \textbf{0.72±0.10} & 0.81±0.07 & - \\
&&&&& OA-5 & 0.44±0.12&0.41±0.13 &  0.42±0.13 &0.77±0.07& 0.73±0.12 \\
\hline

\textbf{\multirow{3}{*}{all rad.}} & \multirow{3}{*}{\~\scriptsize\(10^3\)} & \multirow{3}{*}{-} & \multirow{3}{*}{-} & \multirow{3}{*}{-}
& ACL &0.68±0.08& 0.49±0.10 &  \textbf{0.51±0.09}&  0.63±0.15& 0.29±0.19 \\
&&&&& OA-b & 0.69±0.08& 0.67±0.08 & 0.68±0.07 & 0.81±0.06 & - \\
&&&&& OA-5 &0.43±0.11&0.40±0.11 & 0.45±0.10&   0.73±0.08 & 0.61±0.13 \\
\hline

\textbf{\multirow{6}{*}{Top-k\cite{chen2025patient}}} & \multirow{3}{*}{25} & \multirow{3}{*}{10} & \multirow{3}{*}{1000} & \multirow{3}{*}{3}
& ACL & 0.64±0.02& 0.48±0.03 & 0.53±0.05 & 0.72±0.04 & 0.31±0.07 \\
&&&&& OA-b & 0.61±0.10 & 0.68±0.08& 0.61±0.12& 0.67±0.13&  -\\
&&&&& OA-5 &0.34±0.08 & 0.30±0.11& 0.33±0.10& 0.61±0.05&   0.38±0.11 \\
\cline{2-11}
& \multirow{3}{*}{30} & \multirow{3}{*}{10} & \multirow{3}{*}{1000} & \multirow{3}{*}{3}
& ACL & 0.57±0.02& 0.43±0.03 & 0.50±0.05 & 0.70±0.03 & 0.25±0.06 \\
&&&&& OA-b & 0.70±0.10& 0.75±0.08 &0.69±0.12& 0.77±0.11 &  - \\
&&&&& OA-5 &0.41±0.13& 0.38±0.15 & 0.43±0.16& 0.75±0.10& 0.55±0.22 \\
\hline

\textbf{\multirow{6}{*}{Ours}} & \multirow{3}{*}{25} & \multirow{3}{*}{10} & \multirow{3}{*}{1000} & \multirow{3}{*}{3}
&ACL &0.67±0.04 & 0.51±0.03 &\textbf{0.56±0.11}  & 0.64±0.08 & 0.27±0.13 \\
&&&&& OA-b &\textbf{0.75±0.12}& 0.71±0.14 &0.71±0.14  &0.74±0.06 &- \\
&&&&& OA-5 &\textbf{0.44±0.06}& 0.39±0.04 & \textbf{0.43±0.08} &0.70±0.05  & 0.56±0.12 \\
\cline{2-11}
& \multirow{3}{*}{30} & \multirow{3}{*}{10} & \multirow{3}{*}{1000} & \multirow{3}{*}{3}
& ACL &\textbf{0.73±0.03} & 0.45±0.08 &0.46±0.09  & 0.66±0.02 & 0.30±0.08 \\
&&&&& OA-b &0.72±0.10 &0.70±0.11  &\textbf{0.72±0.12}&0.74±0.13  &  -  \\
&&&&& OA-5 &0.43±0.10& 0.36±0.10& 0.40±0.09&0.66±0.10 & 0.50±0.12 \\
\hline
\end{tabular}
\end{table}
\noindent\textbf{Datasets.}
\textit{\textbf{ACL tear detection.}}
We use the public clinical knee MRI dataset from the Clinical Hospital Centre Rijeka PACS~\cite{vstajduhar2017semi}, consisting of sagittal-plane volumes with clinically confirmed ACL tear labels. We use 664 cases for training and 92 for validation, and perform three-class classification (\emph{non-tear}, \emph{partial tear}, \emph{complete tear}). \textit{\textbf{OAI-ZIB-CM.}} To evaluate generalisability on an osteoarthritis cohort, we use the public OAI-ZIB-CM dataset derived from the Osteoarthritis Initiative (OAI), containing 507 knee MR volumes with manual masks for five anatomical ROIs (Femur, Femoral Cartilage, Tibia, Medial Tibial Cartilage, and Lateral Tibial Cartilage) and radiographic OA severity labels via KL grade. We derive a binary OA label by grouping KL$\in\{0,1\}$ as non-OA and KL$\in\{2,3,4\}$ as OA~\cite{felson2011defining}.
All images are preprocessed and resized to $32\times128\times128$.

\noindent\textbf{ROI definition \& feature extraction.}
For OAI-ZIB-CM, we train nnU-Net on the provided 5-ROI annotations to segment bone and cartilage. For the ACL cohort, a radiologist annotated ACL masks for 45 cases, which we use to train nnU-Net for ACL segmentation.
 On a held-out test split, mean$\pm$std Dice is $0.98\pm0.00$ (Femur), $0.88\pm0.02$ (Femoral cartilage), $0.99\pm0.00$ (Tibia), $0.84\pm0.04$ (Medial tibial cartilage), $0.86\pm0.04$ (Lateral tibial cartilage), and $0.65\pm0.08$ (ACL). For ACL dataset, we additionally define coarse context ROIs by taking a centred crop (50\% depth, 30\% height, 50\% width) and partitioning it into a $2{\times}2{\times}2$ grid, giving 9 ROIs in total.
We extract radiomic features per ROI using PyRadiomics~\cite{van2017computational}: first-order, 3D shape, and texture families (GLCM, GLRLM, GLSZM, NGTDM, GLDM), yielding 107 features per ROI. 

\noindent\textbf{Implementation details}
The scorer $s_{\psi}$ uses an image encoder to extract context from $\mathbf{x}_i$ and a 2-layer MLP to fuse this context with the set embedding into a scalar utility score.
The feature-set encoder $\mathcal{E}_{\theta}$ follows a DeepSets-style design~\cite{zaheer2017deep}, forming permutation-invariant set embeddings by tokenising each selected feature (value + learned metadata embedding), applying an MLP to each token, and mean-pooling across the $k$ tokens.
Both the probe $q_{\phi}$ and the downstream classifier $c_{\eta}$ are lightweight linear classifiers (logistic regression). In \textbf{Stage~1}, we warm-start the scorer $s_{\psi}$ for 80 epochs.
For each subject, we compute probe rewards for 50 randomly sampled feature sets (we also tested 30/70, with no improvement beyond 50).
In \textbf{Stage~2}, we sample a preliminary pool $\mathcal{P}^{(0)}_i$ with $|\mathcal{P}^{(0)}_i|\in\{2000,5000,10000\}$, score all candidates with $s_{\psi}$, and retain the top-$M$ sets to form $\mathcal{P}_i$, with $M\in\{500,1000,2000\}$ (no improvement beyond $|\mathcal{P}^{(0)}_i|=5000$ and $M=1000$). For the training-only probe, we ablate the number of optimisation steps and use 10 steps as a trade-off between performance and cost. At inference, we optionally ensemble the top-3 retrieved feature sets by averaging their logits before softmax.

\noindent\textbf{Comparison and ablation Results.}
Table~\ref{ablation} compares our proposed method with end-to-end image baselines, radiomics using all features, and marginal top-$k$ selection~\cite{chen2025patient}. Our feature-set retrieval is competitive with image baselines while enforcing an auditable $k$-feature evidence set. On ACL, the best setting achieves \textbf{0.73$\pm$0.03} accuracy ($k{=}30$), comparable to Img+msk (0.72$\pm$0.03) and above end-to-end (0.70$\pm$0.05), and it outperforms top-$k$ at matched $k$ with a significant gain in accuracy ($p=0.01$ for Acc./BAcc). On OA-b, we obtain up to \textbf{0.75$\pm$0.12} accuracy (ours, $k{=}25$), exceeding top-$k$ at $k{=}25$ and matching/exceeding image baselines and all-radiomics. On OA-5, our method yields competitive ordinal agreement (QWK up to 0.56$\pm$0.12) with comparable Acc./BAcc. to baselines. Ablations indicate insufficient probe optimisation consistently degrades results, supporting 10 probe steps as a stable trade-off; overly small candidate pools reduce performance; $k$ reflects the expected compactness--coverage trade-off (smaller $k$ under-covers evidence; larger $k$ dilute minority-class discrimination); and ensembling the top retrieved sets provides small but consistent gains.

\noindent\textbf{Interpretations.} Each subject's primary feature set $S_i^\star$ enables per-patient interpretation. \textbf{Case 1 }is a partial ACL tear: although some fibres remain continuous, intraligamentous high signal is visible along the ACL (arrow; Fig.~\ref{fig:case_study}).  About 25\% of influential features come from the ACL ROI, indicating reliance on ligament-specific signal. Within this ROI, first-order Maximum and GLRLM Gray Level Variance contributed strongly, consistent with focal high signal and increased intraligamentous heterogeneity,  typically observed in partial tearing. Additional contributions arose from the medial-anterior and lateral-posterior subpatches, spatially consistent with the ACL course between the tibial origin and femoral insertion \cite{PetersenZantop2007}. \noindent\textbf{Case 2} (non-tear) shows a continuous, normal-appearing ACL. Compared with Case 1, fewer selected features originate from the ACL ROI, with greater contribution from the central region and a more lateral/anterior emphasis. This suggests that without ligament disruption, the model relies more on contextual texture organisation than ACL-centred heterogeneity. Cases 1–2 indicate that top-ranked features capture either lesion-specific ACL signal or indirect contextual cues linked to ACL status. \noindent\textbf{Case 3} is a severe knee OA case~\cite{KellgrenLawrence1957}, showing extensive cartilage thinning(yellow arrows), joint-space loss, and bony irregularity (red arrow). 10/30 selected features come from the femoral cartilage. Leading features (Skewness, GLCM Correlation/Energy, GLDM Dependence Non-Uniformity) reflect altered intensity distribution, texture correlation, and heterogeneity in cartilage and subchondral regions, consistent with advanced OA degeneration and subchondral remodelling~\cite{Pane2023}.

\section{Discussion and Conclusion}

Our work extends adaptive radiomics beyond marginal top-\(k\) weighting to patient-specific feature-set selection, predicting a single compact \(k\)-feature set per subject to optimally capture complementary evidence across radiomic families and anatomical ROIs. We introduce a two-stage retrieval procedure that approximates exhaustive \(\binom{F}{k}\) search with a small per-subject candidate pool. Across ACL tear detection and KL grading, the proposed framework consistently outperforms top-\(k\) selection and remains competitive with baselines, both of which may be considered too large to be practically interpretable. Overall, the proposed retrieval framework offers one of the first practical solutions to transparent, patient-specific radiomics with optimality-bounded evidence coverage.

\section{Acknowledgement}
This work was supported by the International Alliance for Cancer Early Detection (an alliance between Cancer Research UK [EDDAPA-2024/100014] and [C73666/A31378], Canary Center at Stanford University, the University of Cambridge, OHSU Knight Cancer Institute, University College London, and the University of Manchester) and the National Institute for Health Research University College London Hospitals Biomedical Research Centre, and was also supported by the European Union’s Horizon Europe research and innovation programme under the Marie Skłodowska-Curie Actions Doctoral Networks (grant agreement No. 101227121—RENOVATE; HORIZON-MSCA-2024-DN-01).

\section{Disclosure of Interests}
The authors declare that they have no conflicts of interest related to this work.

\bibliographystyle{splncs04}
\bibliography{Paper-86}

\end{document}